%% file: icdm.tex
\documentclass[conference]{IEEEtran}

\IEEEoverridecommandlockouts

\usepackage{times}
\usepackage{helvet}
\usepackage{courier}
\usepackage{textcomp}

\usepackage{psfig}
\usepackage{subfig}
\usepackage[linesnumbered,ruled,vlined]{algorithm2e}
\usepackage{floatflt}
\usepackage{rotating}
\usepackage{multicol}
\usepackage{multirow}
\usepackage{color}
\usepackage{epsfig}
\usepackage{amsmath}
\usepackage{amsfonts}
\usepackage{array}
\usepackage{flushend}

\newcommand{\RNum}[1]{\uppercase\expandafter{\romannumeral #1\relax}}

\begin{document}

\title{Layerwise Perturbation-Based Adversarial Training for Hard Drive Health Degree Prediction
}


\author{\IEEEauthorblockN{Jianguo Zhang\IEEEauthorrefmark{1},
Ji Wang\thanks{Jianguo Zhang and Ji Wang contributed equally to this work. Lifang He is the corresponding author.}\IEEEauthorrefmark{2}, 
Lifang He\IEEEauthorrefmark{3},
Zhao Li\IEEEauthorrefmark{4}, 
Philip S. Yu\IEEEauthorrefmark{1}\IEEEauthorrefmark{5}
}
\IEEEauthorblockA{\IEEEauthorrefmark{1}Department of Computer Science, University of Illinois at Chicago, Chicago,
IL, USA; \{jzhan51, psyu\}@uic.edu}
\IEEEauthorblockA{\IEEEauthorrefmark{2}College of Systems Engineering,
National University of Defense Technology, Changsha, Hunan, China; wangji@nudt.edu.cn} 
\IEEEauthorblockA{\IEEEauthorrefmark{3}Weill Cornell Department of Healthcare Policy \& Research, Cornell University, NY, USA; lifanghescut@gmail.com}
\IEEEauthorblockA{\IEEEauthorrefmark{4}Alibaba Group, Hangzhou, Zhejiang, China; lizhao.lz@alibaba-inc.com }
\IEEEauthorblockA{\IEEEauthorrefmark{5}
Shanghai Institute for Advanced Communication and Data Science, \\ 
Shanghai Key Laboratory of Data Science, Fudan University, Shanghai, China}
}



\maketitle

\begin{abstract}
With the development of cloud computing and big data, the reliability of data storage systems becomes increasingly important. Previous researchers have shown that machine learning algorithms based on SMART attributes are effective methods to predict hard drive failures. In this paper, we use SMART attributes to predict hard drive health degrees which are helpful for taking different fault tolerant actions in advance. Given the highly imbalanced SMART datasets, it is a nontrivial work to predict the health degree precisely. The proposed model would encounter overfitting and biased fitting problems if it is trained by the traditional methods. In order to resolve this problem, we propose two strategies to better utilize imbalanced data and improve performance. Firstly, we design a layerwise perturbation-based adversarial training method which can add perturbations to any layers of a neural network to improve the generalization of the network. Secondly, we extend the training method to the semi-supervised settings. Then, it is possible to utilize unlabeled data that have a potential of failure to further improve the performance of the model.
Our extensive experiments on two real-world hard drive datasets demonstrate the superiority of the proposed schemes for both supervised and semi-supervised classification. The model trained by the proposed method can correctly predict the hard drive health status 5 and 15 days in advance. Finally, we verify the generality of the proposed training method in other similar anomaly detection tasks where the dataset is imbalanced. The results argue that the proposed methods are applicable to other domains.
\end{abstract}


\section{Introduction}
Nowadays, increasing numbers of industrial and academic institutes rely on data centers to store and process their data. The crash of data centers may incur tremendous loss or even catastrophic consequences. The reliability and the availability of data centers are of the utmost importance to data center administrators. However, the complex architecture and functionality of data centers lead to a serious problem of IT equipment failures, among which hard drives are the most frequently failing components \cite{Sankar2013,Botezatu2016}. Hence, it is in high demand to take measures to handle the hard drive failure issue.

The self-monitoring, analysis and reporting technology (SMART) \cite{allen2004monitoring} has been implemented in almost all hard drives to monitor and analyze the internal attributes of hard drives. The previous researches demonstrate that the impending hard drives failure manifests itself through SMART statistics \cite{backblaze2015}. It is feasible to predict the impending failures by using SMART statistics. To improve failure prediction performance, many efforts have been made based on SMART attributes, including analyzing the failure behaviors of hard drives \cite{hughes2002improved,Botezatu2016}, and designing machine learning algorithms for predicting hard drive failures \cite{murray2005machine,zhu2013proactive,Mahdisoltani2017}. Most of these works focused on the proactive failure prediction, which forecasts hard drive failures in advance and gives a binary result identifying the hard drive as healthy or faulted. However, hard drives usually fail gradually rather than abruptly. To harness the potential of gradual change, it is necessary to develop a health prediction model which can predict the heath status of hard drives rather than merely providing a simple binary result.

Predicting the health status of hard drives is not a trivial task. Most hard drives undergo a deterioration process before they finally fail. The SMART statistics begin to accumulate deviation from the normal state days before the final failure. Hence, we need to extract the long-term temporal dependency in the SMART statistics to make an accurate prediction of the hard drive status. The most stubborn problem in the prediction of health status is that the data to be used in the model is highly imbalanced. Although hard drive failures occur frequently in data centers, the failure records are much fewer than the healthy records, merely accounting for less than 3\% of the total records \cite{backblaze2015}. Due to the overfitting and biased fitting problems of most statistical and machine learning algorithms, the prediction model trained by the imbalanced SAMRT data is easy to be biased fitted to the healthy records and over fitted to the failure records, which results in a poor predictive performance. Some novel methods are desirable to tackle imbalance issue and establish a high quality prediction model by using the SMART statistics.

In order to extract the long-term temporal dependency embedded in the SMART statistics, we build a deep neural network based on the long short-term memory unit (LSTM) \cite{Hochreiter1997} which specializes in processing sequential data \cite{Graves2013,sun2017sequential}. Nonetheless, the deep neural networks (DNNs) are notorious for the overfitting and biased fitting problems on the highly imbalanced datasets like the hard drive records used in this work. To solve this problem, we propose two strategies in this paper.

Firstly, we introduce the adversarial training strategy. The adversarial training \cite{szegedy2013intriguing,goodfellow2014explaining} is a strong regularization method that injects the perturbations affecting the neural network's inference in the most sensitive way during the training phase. The traditional adversarial training injects adversarial perturbations into the inputs to force the networks to learn a better distribution of the training data and avoid the overfitting problem \cite{miyato2016adversarial,kurakin2017adversarial}. However, only injecting perturbations into the inputs may restrict the effectiveness of the adversarial training. In this paper, we enhance the adversarial training by enabling layerwise perturbation where the adversarial perturbation can be injected into any layer of the neural network rather than just the input layer.

Another more effective approach to mitigate the overfitting and biased fitting is to increase the number of positive samples, i.e., the failure records. It is obvious that the anomaly of SMART statistics is evident when hard drives are close to failure. Therefore, we can label the records close to failure as the failure records directly. But for the records dozens of days ahead of the final failure, they may be less informative for the failure or even demonstrate the same features as the healthy records do. It is inappropriate to label these records as failure records arbitrarily. Labeling these records is a time consuming task and relies on the expert knowledge, which makes it unfeasible in reality. Hence, in traditional supervised training, these records are discarded in spite of the high probability that they show fault features. To fully harness these potential failure records, we extend the proposed adversarial training into the semi-supervised setting where the records far before the final failure are regarded as unlabeled data during the training phase.

Based on the above two strategies, we propose a \textbf{L}ayerwise \textbf{P}erturbation-based \textbf{A}dversarial \textbf{T}raining (LPAT) method to train our hard drive status prediction model. Note that although the LPAT is designed for predicting hard drive health status in this work, it can be used in other similar anomaly detection problems to mitigate the overfitting and biased fitting. The main contributions are summarized as follows:
\begin{itemize}
\item We design a novel LPAT method for hard drive status prediction tasks. Instead of only adding perturbation to inputs of neural networks, LPAT can flexibly add perturbation to a specific layer or all layers to better address the overfitting and biased fitting problem. We further introduce an approach based on Kullback-Leibler divergence to utilize unlabeled data to improve the performance of the prediction model.
\item To the best of our knowledge, it is the first work toward designing a layerwise perturbation-based adversarial training with the semi-supervised setting for deep learning models. We use this model to predict hard drive status in both supervised and semi-supervised settings.
\item Thorough experiments on two hard drive datasets demonstrate that LPAT can improve the prediction performance in both supervised and semi-supervised settings. Specifically, the model can predict hard drive health status 5 days and 15 days before failure by using sequential SMART statistics, which is more useful in practice than only predicting healthy or failed. In addition, we apply the proposed methods to a image recognition task and a sequential analysis task to verify their generality.
\end{itemize}

The rest of the paper is organized as follows. We define the prediction problem in Section \RNum{2}. Then, Section \RNum{3} presents the proposed methodology. In Section \RNum{4}, we evaluate our methods thoroughly by conducting groups of experiments. Section \RNum{5} summarizes related work. Finally, Section \RNum{6} concludes this paper and describes the future work.

\section{Problem Definition}
In this work, we aim to build a model trained by LPAT to predict the status of hard drives based on their SMART statistics. The training dataset is denoted as $D=\left \{ (x^{(1)}, y^{(1)}),..., (x^{(N)}, y^{(N)})\right \}$, where $x^{(i)}\in R^{w\times n}$ is the $i^{th}$ training sample consisting of continuous SMART records from the day $t_{i}$ to the day $t_{i}+w$. Each day has a feature vector of $n$ SMART attributes, and $y$ denotes health degrees of the hard drive. We predict hard drives by three different health degrees\footnote{Note that we choose the health degree intuitively; other definitions can be used with only a slight modification.}, i.e., $ y \in \left \{0,1,2\right\}$. `0' represents a ``red alert" which means the residual life of the hard drive is less than $5$ days; `1' represents that the drive is ``going to fail" in $5-15$ days; `2' means ``healthy". Our goal is to learn a function $f: X\rightarrow \left \{ 0,1,2 \right \}$ that minimizes the negative log-likelihood $\hat{\mathcal{L}}(\Theta)$ on the training dataset, where $\Theta$ is the model parameters learned during the training. Intuitively, we try to train a model that correctly predicts the health degree of a hard drive. $\hat{\mathcal{L}}(\Theta)$ can be formulated as: 
\begin{equation}\label{basic-loss}
\hat{\mathcal{L}}(\Theta)=-\frac{1}{N}\sum_{i=1}^{N}log\ p(y^{(i)}|x^{(i)}; \Theta ),
\end{equation}

\section{Methodology}
\subsection{LSTM Based Neural Network}
\begin{figure}[t]
    \centering
    \includegraphics[width=0.65\linewidth]{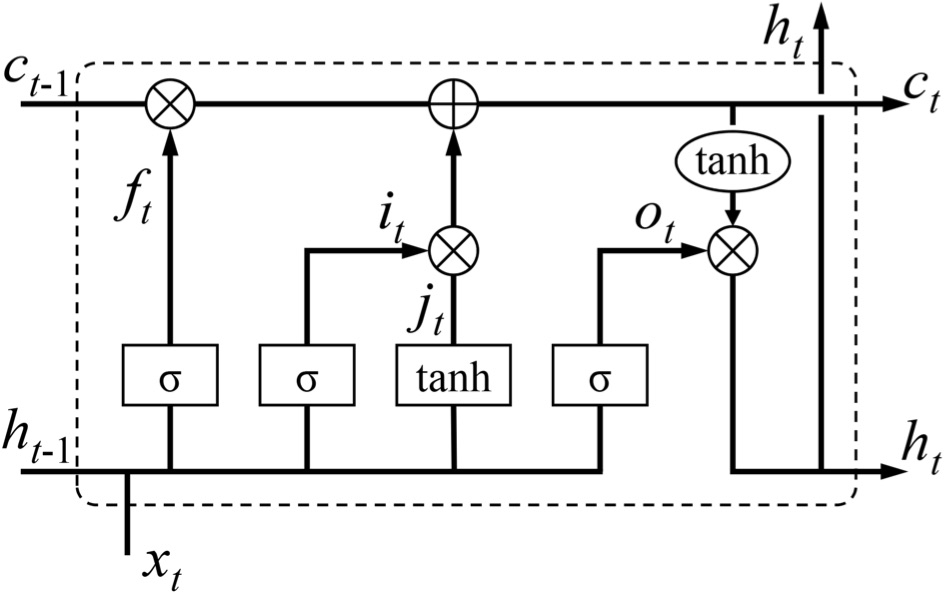}
    \caption{A component of the LSTM layer.}
    \label{fig:lstm}
\end{figure}

The proposed prediction model includes two dense (fully connected) layers, followed by a LSTM layer and a dense layer. LSTM is the basic building block used to extract the long-term temporal dependency in sequential SMART statistics. Fig. \ref{fig:lstm} is an illustration of the component in the LSTM layer. It is formulated as:
\begin{equation}
\hat{\alpha}_{t}=W_{\alpha}x_{t}+U_{\alpha}h_{t-1}+b_{\alpha},
\end{equation}
\begin{equation}
c_{t}=i_{t}\odot j_{t} + f_{t}\odot c_{t-1},
\end{equation}
\begin{equation}
h_{t}=o_{t}\odot tanh(c_{t}),
\end{equation}
where $\alpha \in \{  i, f,o,j \}$. $i_t,f_t,o_t$ are the $input\ gate$, $forget\ gate$ and $output\ gate$, respectively. $c_{t}$ represents the \textit{memory cell} which is designed to counteract the problem of vanishing/exploding gradient, and thus enable extracting long-term temporal dependency. The $forget\ gate$ is for resetting the memory cells. The $input\ gate$ and the $output\ gate$ are for controlling the input and the output of the memory cells. $\alpha_{t}=\sigma(\hat{\alpha}_{t})$ when $\alpha \in \{i,f,o\}$, where $\sigma$ is the sigmoid activation function. $j_{t}=\tanh (\hat{j_{t}})$ is a proposed update to the cell state. $x_{t}\in R^{n}$ is the input from the lower dense layer at the time step $t$. $h_t$ is the output of current block at the time step $t$. In the $q$ LSTM units, the weight matrices $W_{\alpha }\in R^{q\times d}$, $U_{\alpha }\in R^{q\times d}$ and the bias vectors $b_{\alpha }\in R^{q}$ are the parameters to learn for $\alpha  \in \{i,f,o,j\}$.
\subsection{Layerwise Perturbation-Based Adversarial Training}
\begin{figure}
    \centering
    \includegraphics[width=1.0\linewidth]{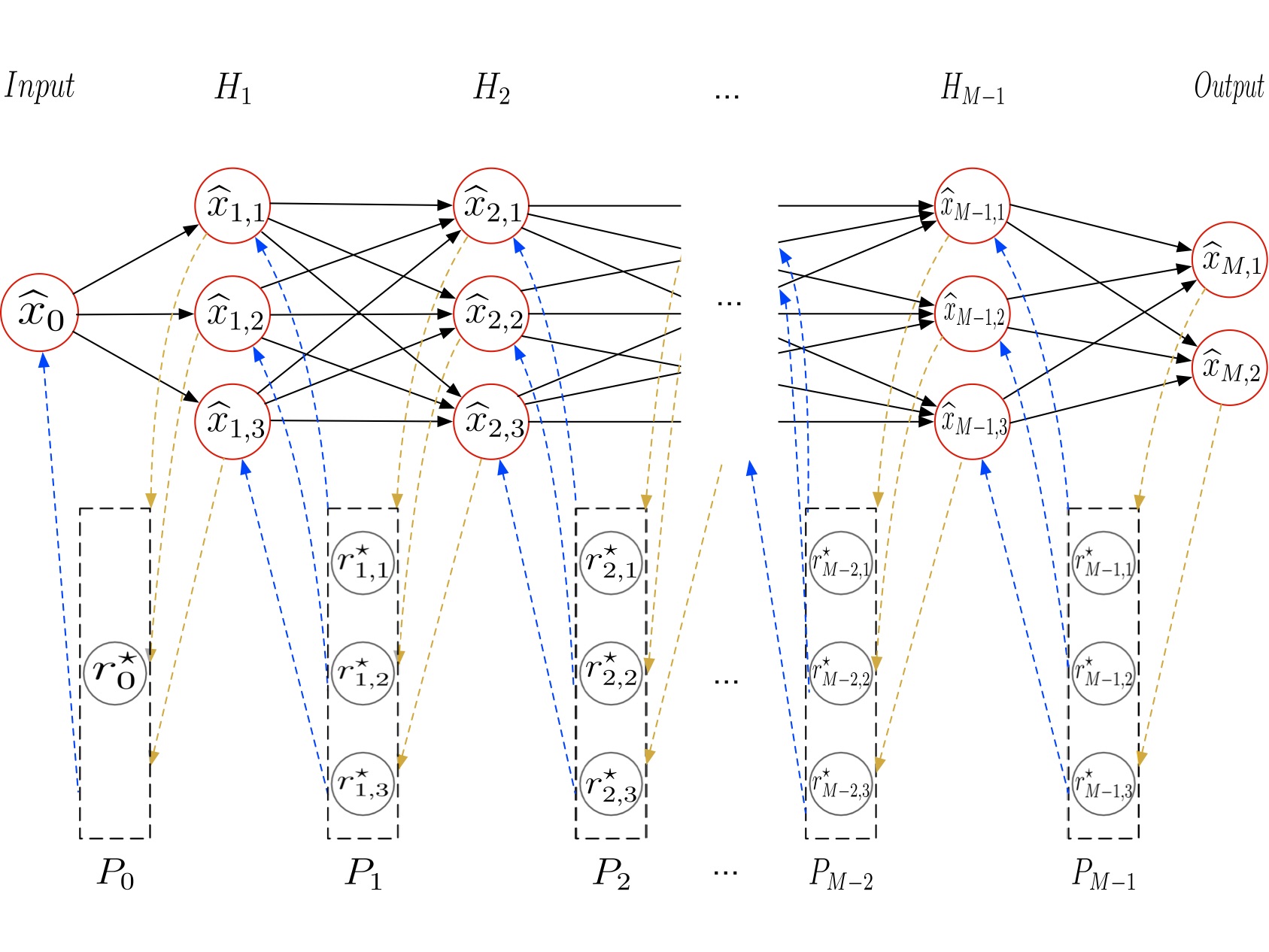}
    \caption{An overview of LPAT. It goes through two rounds of feedforward and backpropagation. In the first round, it performs feedforward process to compute the output of the neural network. Then, it performs backpropagation to update parameters and store gradients in gradient accumulation layers. In the second round, it adds layerwise adversarial perturbations based on the gradients to each layer and performs feedforward process to compute the new output of the neural network. Finally, it performs backpropagation again to update parameters of all network layers.}
    \label{fig-framework}
\end{figure}

Previous works on adversarial training concluded that training a DNN with adversarial examples acts as a regularizer and improves the robustness of the neural network on the test dataset. In order to mitigate the overfitting and biased fitting problem caused by the highly imbalanced data, we design LPAT to train the prediction model. Instead of only injecting perturbation into inputs, LPAT generates adversarial samples at the time series inputs and the intermediate layers. Fig. \ref{fig-framework} illustrates how LPAT works when training the model.

In Fig. \ref{fig-framework}, for a model consisting of $M-1$ hidden layers, $m=0$ is the input layer, and $\hat{x}_{m}=(\hat{x}_{m,1},...,\hat{x}_{m,k})$ is the output of the $m^{th}$ layer, where $k$ is the dimension of the layer. Each layer has a gradient accumulation layer $P_{m}$ serving two functions: (1) it temporarily stores the backpropagation gradients on the output of the $m^{th}$ layer, which is denoted by the yellow line; and (2) it computes the adversarial perturbations $r_{m}^{\star}=(r_{m,1}^{\star},...,r_{m,k}^{\star})$ for the $m^{th}$ layer and adds the perturbation to $\hat{x}_{m}$ as denoted by the blue line. Then, the neural network performs feedforward process again to compute its new output. The training process can be formed as a min-max problem. The adversarial samples apply the worst perturbation to maximize the error of the model, while the model tries to be robust to such perturbations through minimizing the error caused by the adversary. The min-max problem for the $m^{th}$ layer can be formulated as an additional cost function:
\begin{equation}
\underset{\Theta}{min} \underset{r_m, \left \| r_m \right \|\leq \epsilon }{max} l(p(y^{(i)}| \hat{x}_m^{(i)}+r_m,\Theta), q(y^{(i)}))
\end{equation}
where $\epsilon$ is the magnitude of the perturbation, $l$ is the non-negative cross entropy between two distributions representing the prediction error caused by the perturbation, and $q(y^{(i)})$ is a one-hot distribution of the corresponding label $y^{(i)}$. The additional cost function can be simplified as: 
\begin{equation}\label{equ:adv_loss}
-log\ p(y^{(i)}| \hat{x}_m^{(i)}+r_{m};\Theta).
\end{equation}
Then, 
\begin{equation}
r^{\star}_{m}= arg \min\limits_{r_m,\left \| r_m \right \|\leq \epsilon}\ log\ p(y^{(i)}| \hat{x}^{(i)}+r_m;\hat{\Theta}),
\end{equation}
where $\hat{\Theta}$ is a constant set to the current parameters of the model. In general, the exact calculation for $r_m$ is intractable for most DNNs. We follow the method proposed in \cite{goodfellow2014explaining} to linearly approximate $r_m$ with the $L_{2}$ norm constraints as:
\begin{equation} \label{eqn:rstar}
r^{\star}_{m}\approx -\epsilon\frac{g}{\left \| g \right \|_{2}},
\end{equation}
where $g=\nabla _{ \hat{x}_m^{(i)}} log\ p(y^{(i)}| \hat{x}_m^{(i)};\hat{\Theta}).$

\begin{algorithm*}[t]
\SetAlgoVlined
\small
\KwIn{Randomly initialized Network $NN$. $B^i$ is the batch sampled at step $i$ of size $k$, with labeled training samples $(X_l^i, Y_l^i)$ and unlabeled training samples $(X_{ul}^i, \cdot)$. $\left \{ P_{i} \right \}_{i=0}^{P-1}$ are the gradient accumulation layers with stored layerwise adversarial perturbations $ \left \{r^{\star }_{i}\right \}_{i=0}^{P-1}$ initialized with zero. No gradient accumulation layer is active at the initial step $i=0$. $\epsilon_{m}$ denotes perturbation magnitude for layer $m$.}

\For{$i\gets0$ \KwTo $B-1$}{
    Sample a batch consisting of $(X_l^i, Y_l^i)$ and $(X_{ul}^i, \cdot)$ of size $k$ from training dataset\;
    Perform feedforward process without perturbation to calculate the output of $NN$ and the loss $\hat{\mathcal{L}}(\Theta )$ for $(X_l^i, Y_l^i)$ using Eq. (\ref{basic-loss})\;
    Produce a random unit vector $e$ using an iid Gaussian distribution\;
    Perform backpropagation to take the gradient of $KL$ w.r.t $r$ on $r = \xi e$ on each layer's output $\hat{X}_{m}$. Each accumulation layer $P_{m}$ temporarily stores the gradients backpropagated to that layer\;
    Calculate and store layerwise adversarial perturbation $r^{\star }_{m}$ for each layer using  Eq. $(\ref{rvat})$\;
    Perform feedforward process with layerwise adversarial perturbation: $\hat{X}_{m}=\hat{X}_{m}+\epsilon_{m} \cdot r^{\star }_{m}$. Calculate the loss $\mathcal{L}_{lap}$ for both $(X_l^i, Y_l^i)$ and $(X_{ul}^i, \cdot)$ using Eq. (\ref{eqn:loss-lap})\;
    Perform backpropagation using loss ${\mathcal{L}}(\Theta)$ to update parameters\;
}
\small\caption{Layerwise Perturbation-Based Adversarial Training}
\label{alg:1}
\end{algorithm*}

\subsection{Extension of Semi-Supervised Setting}
We extend the proposed adversarial training into the semi-supervised setting to utilize the records relatively far before the final failure. Due to the difficulty of labeling these records, we regard them as unlabeled samples during the training. Then, the adversarial perturbation $r^{\star}$ is incalculable by using Eq. (\ref{eqn:rstar}) without the label $y^{(i)}$. To depict the prediction error, we introduce the Kullback-Leibler (KL) divergence \cite{Kullback1951} which measures the divergence between the current output distribution $p(\cdot |x);\hat{\Theta })$ and the perturbed output distribution $p(\cdot |x+r;\hat{\Theta } )$.  The layerwiase adversarial perturbation for the $m^{th}$ layer can be calculated as:
\begin{equation}\label{layer-wise-asv}
r^{\star }_{m}=arg \max \limits_{r_{m}, \left \| r_{m} \right \|\leq \epsilon } KL_{m},
\end{equation}
where
\begin{equation}\label{layer-wise-asv1}
    KL_{m}=KL[ p(\cdot |\hat{x}_{m};\hat{\Theta}) \left |  \right |p(\cdot |\hat{x}_{m}+r_{m};\hat\Theta )].
\end{equation}
$p(\cdot |\hat{x}_{m};\hat{\Theta})$ is differential with $\hat{\Theta}$ and $\hat{x}_{m}$. However, the maximum value of $KL_{m}$ cannot be computed directly as the first derivative $\nabla_{r_{m}} KL_{m}$ has the minimum value $0$ when $r_{m}=0$. Hence, we approximate it with the second-order Taylor series:
\begin{equation}
K{L_m} \approx \frac{1}{2}{r_m}^TH(\hat{x}_{m},\hat{\Theta}){r_m},
\end{equation}
where $H(\hat{x}_{m},\hat{\Theta})$ is the Hessian matrix \cite{golub2000eigenvalue}. Then, $r_m^{\star}$ is the first dominant eigenvector $u(\hat{x}_{m},\hat{\Theta})$ of $H(\hat{x}_{m},\hat{\Theta})$ with magnitude $\epsilon$, and $r_m^{\star}$ can be represented as:
\begin{equation}
r_m^{\star} = \epsilon \overline {u(\hat{x}_{m},\hat{\Theta})},    
\end{equation}
 where $\overline{(\cdot)}$ represents the unit vector in the direction of $(\cdot)$. Given a randomly sampled unit vector $e$, according to the power iteration method \cite{golub2000eigenvalue}, $u(\hat{x}_{m},\hat{\Theta})$ is the convergence of the iteration $e \leftarrow \overline {He}$, and $He$ can be approximated as follow:
\begin{equation}
He \approx \frac{{{\nabla _r}K{L_m}{|_{r = \xi e}} - {\nabla _r}K{L_m}{|_{r = 0}}}}{\xi },
\end{equation}
where $\xi\ne 0$. The preliminary test shows that one-time iteration provides sufficient performance as multi-time iterations in our problem. So we approximate the convergence by the one-time power iteration \cite{golub2000eigenvalue}. Then, the approximation of $r_m^{\star}$ is:
\begin{equation}\label{rvat}
 r^{\star }_{m}\approx \epsilon \frac{g}{\left \| g \right \|_{2}},
\end{equation}
where
\begin{equation}
    g=\nabla _{r}KL[p(\cdot |\hat{x}_{m};\hat{\Theta }) \left |  \right |p(\cdot |\hat{x}_{m}+r;\hat\Theta )] |_{r = \xi e}.
\end{equation}

Hitherto, we get the adversarial perturbation to the DNN model in both the supervised setting and the semi-supervised setting. During the training phase, the DNN model is trained not only to minimize the regular classification loss on labeled data but also to resist the adversarial perturbation on labeled and unlabeled data. Thus, for all the training dataset including labeled and unlabeled data of size $N'$, the full loss is given by:
\begin{equation}
\mathcal{L}(\Theta ) = \hat{\mathcal{L}}(\Theta )+\lambda \cdot \mathcal{L}_{lap},
\end{equation}
where $\hat{\mathcal{L}}(\Theta )$ is the negative log-likelihood for the labeled data defined by Eq. $(\ref{basic-loss})$. $\mathcal{L}_{lap}$ is computed on both labeled and unlabeled data:
\begin{equation}
\label{eqn:loss-lap}
\mathcal{L}_{lap}=\frac{1}{N'}\sum_{i=1}^{N'} KL[  p(\cdot |\hat{x}^{(i)}_{m};\hat\Theta_{m}) \left |  \right |p(\cdot |\hat{x}^{(i)}_{m}+{r^{ \star}_{m}}^{(i)};\hat\Theta_{m} )],
\end{equation}
for $m=0, 1, ..., M-1$.

The gradients computed from mini-batch inputs aggregated through intermediate layers are used to calculate perturbations for the current mini-batch. Since Eq. $(\ref{rvat})$ requires no label information, it is not necessary to shuffle mini-batch data to avoid overlap of labels. Note that the number of accumulation layers $P$ is not necessarily equal to the number of model layers as we can add perturbations to any layers. When $P=0$, we only inject perturbations into the input. $P=M-1$ indicates that all layers of the model are perturbed. The training procedure is summarized in Algorithm $\ref{alg:1}$. Parameter $\epsilon_{m}$ can be set to a fixed value for all layers or set differently for different layers. During training, we add the gradient accumulation layer $P_{m}$ after each neural network layer. 
During testing, all the gradient accumulation layers are removed from the model.

\input{experiment.tex}

\section{Related Work}\label{sec:relatedwork}
A series of models have been proposed to predict the impending failures of hard drives. A nonparametric model was proposed by \cite{Hughes2002} to detect the anomalous SMART values of hard drives. \cite{Hamerly2001} designed a Bayesian classifier to predict the hard drive failures. In order to capture the temporal features embedded in the SMART statistics, some researchers resorted to the time series analysis methods. \cite{Zhao2010} employed the hidden Markov model to predict the imminent failures. \cite{Botezatu2016} used the Bayes-based model to detect the changepoint of SMART statistics, and then compacted the time sequence representation by using an exponential smoothing technique. Besides, the down sampling technology was implemented in \cite{Botezatu2016} to solve the imbalanced dataset problem. All the above works merely attempted to predict the failure, and cannot provide further information about the status of hard drives. \cite{xu2016health} tried to assess the health status of hard drives based on their residual lives. They designed a simple recurrent neural network to classify hard drives as different groups. According to the health status, different protective measures can be taken to improve the storage system reliability. However, the accuracy of their prediction is relatively poor due to the simpleness of their model.

Adversarial training is a strong regularization method which was originally introduced in \cite{goodfellow2014explaining, szegedy2013intriguing}. Their works showed that several deep neural networks are vulnerable to a very small perturbation in the direction that the model's assignment of labels to an unseen class in the most adversarial(sensitive) way. This small perturbation is called adversarial perturbation, which has shown a better performance \cite{miyato2015distributional, wang2017adversary} than dropout \cite{Srivastava:2014:DSW:2627435.2670313} and models trained by random perturbations \cite{goodfellow2014explaining, miyato2015distributional} like adding Gaussian noise \cite{bachman2014learning}. They also found that training the models to be robust against adversarial perturbations was effective to improve performance on testing dataset. \cite{miyato2015distributional} proposed the virtual adversarial method which expanded adversarial training method into semi-supervised learning areas, which improved robustness of models by utilizing the model's posterior distribution against local perturbations around each input data point. They also demonstrated the effects in image classification tasks \cite{kurakin2017adversarial} and text domain \cite{miyato2016adversarial}.  
\cite{wang2018not} designed noise training methods to improve performance of shallow neural networks in mobile cloud.  
\cite{sankaranarayanan2017regularizing} designed methods to improve robustness and performance of very deep neural networks such as VGGnet \cite{simonyan2015very}, InceptionV3 \cite{szegedy2016rethinking} by perturbing intermediate layers. However, above methods either can only be used in supervised learning domains or can only add adversarial perturbations to inputs in semi-supervised learning way.

\section{Conclusion}\label{sec:conclusion}
In this paper, we propose a layerwise perturbation-based adversarial training method with an application on hard drive health degree prediction. Differing from traditional methods which usually predict hard drive health in a binary fashion, the proposed method focuses on different health degrees. Additionally, the proposed LPAT can not only add adversarial perturbations to the input but also to the intermediate layers or all layers, which improves the generalization and performance of model. Besides, it utilizes unlabeled data to further improve performance on prediction. Extensive experiments demonstrate the superiority of our proposed methods.

In the future, we will explore improving performance on more healthy degrees which can give better instructions for technicians and users to take different actions. Also, our method calculates adversarial perturbations each time based on current mini-batch, which is time consuming. We will further optimize the method to reduce computation time and improve performance. In addition, we will extend the adversarial perturbation methods into more other domains which have large-scale unlabeled data such as text classification.

\bibliographystyle{ieeetr}
\bibliography{reference}

\end{document}

%% file: experiment.tex
\section{Experiments}
In order to empirically evaluate the effectiveness of the proposed methods in addressing hard drive healthy degree prediction, we conduct a series of experiments on two datasets and compare it with several existing methods. Furthermore, a real mobile application is used as an experimental example to verify that LPAT can be applied to other similar anomaly detection problems where the dataset is highly imbalanced. In the following, we first introduce the preparation of the experiments. Then we present the experimental results along with the analysis. 

\subsection{Data Preparation}

\begin{table}[t]
\centering
\caption{Number of Healthy (H) and Failed (F) hard drives}
\label{tab:dataset}
\begin{tabular}{|l|l|l|l|l|}
\hline
\multirow{2}{*}{} & \multicolumn{2}{c|}{Original} & \multicolumn{2}{c|}{Post-Processing} \\ \cline{2-5} 
                  & \# of H       & \# of F       & \# of H          & \# of F          \\ \hline
ST-1         & 33800         & 938           & 30685            & 758              \\ \hline
ST-2         & 8660          & 48            &     7932             &      47            \\ \hline
\end{tabular}
\end{table}

Our evaluation and analysis are based on the Backblaze dataset\footnote{https://www.backblaze.com/hard-drive-test-data.html}. At the end of 2016, it recorded 73,653 spinning hard drives. Each entity in the dataset includes the date, the serial number of the hard drive, the model of the hard drive, the SMART statistics, the status, i.e., failed or alive, and other necessary information. 

In this paper, we evaluate the proposed methods based on the SMART statistics of two different models of hard drives in year 2016. The first one is Seagate ST4000DM000 (ST-1). There are 33,800 healthy hard drives and 938 failed hard drives of ST-1 in the dataset. After data cleaning and aggregation \cite{Botezatu2016}, there are 30,685 healthy hard drives and 758 failed hard drives. To further verify our method, we collect another small dataset from Seagate ST8000DM002 (ST-2). There are 8660 healthy hard drives and 48 failed hard drives. After pre-processing there are 7932 healthy drives and 47 failed drives. It should be noted that ST-1 and ST-2 are different models of hard drives although they are from the same hard drive manufacture \cite{Botezatu2016}, which means that they show different degradation progressions. Table \ref{tab:dataset} lists the details.

Apparently, the dataset is extremely imbalanced. For example, there are only 2.47\% samples that are failed in ST-1. The prediction model encounters serious overfitting and biased fitting problems if the dataset is used directly, even with the help of LPAT. Hence, we select the representative subset of healthy hard drives to construct the dataset for training and testing our model, by which the amount of healthy hard drives can be reduced without a significant loss of information about healthy hard drives. The $K$-means clustering algorithm \cite{kanungo2002efficient} is used to cluster the healthy hard drives into 10 clusters based on every day's records of SMART attributes. Then, for each cluster, the top 30\% samples closest to the centroid are selected as the representative subset of healthy hard drives. 

We follow the previous works \cite{murray2005machine,Botezatu2016,xu2016health} to select distinctive SMART features. As the values of different SMART attributes vary widely, we rescale the values of each selected SMART attributes by the following formula to avoid bias to SMART attributes with large values:
\begin{equation}
v' = \frac{{v - {v_{min}}}}{{{v_{max}} - {v_{min}}}},
\end{equation}
where $v$ is the original value of a SMART attribute, and $v_{min}$ and $v_{max}$ are the minimum value and the maximum value of a SMART attribute.

Similar to the definitions used in \cite{li2016being,xu2016health}, we predict the hard drives based on their residual life. Label `0' is ``red alert", which means the residual life is less than 5 days. Label `1' means ``going to fail" in 5 to 15 days. Label `2' represents ``healthy". For the failed hard drives with more than 15-day residual life, we regard them as unlabeled data.

We split each dataset into three subsets, i.e., the training set, the validation set, and testing set. In the supervised setting, we use 80\% of all labeled data as the training set and the rest as the testing set. For ST-1, we use 20\% of the training set as validation set for tuning hyper-parameters. For ST-2, we do not further split the training set due to its small population. In the semi-supervised setting, we add unlabeled data into the training set and keep the validation set and the testing set unchanged. The information about the datasets is summarized in Table \ref{data: split}.

\begin{table}[t]
\centering
\caption{Summary of datasets}
\label{data: split}
\begin{tabular}{|l|c|c|c|c|}
\hline
          & Train & Valid & Test & Unlabeled \\ \hline
ST-1 & 17137 & 4285  & 5356 & 10326      \\ \hline
ST-2 & 755   &  -  & 236  & 489      \\ \hline
\end{tabular}
\end{table}

\begin{table*}[t]
\centering
\caption{Overall results in supervised setting}
\label{tab:ST-1-results}
\begin{tabular}{|p{0.09\textwidth}|p{0.0765\textwidth}<{\centering}|p{0.0765\textwidth}<{\centering}|p{0.0765\textwidth}<{\centering}|p{0.0765\textwidth}<{\centering}|p{0.0765\textwidth}<{\centering}|p{0.0765\textwidth}<{\centering}|p{0.0765\textwidth}<{\centering}|p{0.0765\textwidth}<{\centering}|}
\hline
            & \multicolumn{4}{c|}{ST-1}                & \multicolumn{4}{c|}{ST-2}                \\ \cline{2-9}
            & Accuracy & Precision & Recall & Macro-F1 & Accuracy & Precision & Recall & Macro-F1 \\ \hline
DT          & 82.4     & 73.6      & 74.7   & 74.1     & 82.2     & 74.0      & 72.6   & 73.1     \\ \hline
RGF         & 74.4     & 68.4      & 54.8   & 56.1     & 92.0     & 87.0      & 83.3   & 84.9     \\ \hline
RNN         & 84.3     & 80.1      & 79.8   & 79.9     & 87.7     & 83.9      & 78.6   & 80.9     \\ \hline
basic       & 82.1     & 80.3      & 79.9   & 80.2     & 90.5     & 85.0      & 83.5   & 83.5     \\ \hline
basic+AT    & 86.8     & 81.8      & 81.3   & 81.5     & 91.0     & 85.2      & 84.5   & 84.8     \\ \hline
basic+RDAT    &   88.2   &  83.4     & 83.2  & 83.3     &    92.5  &     88.3  &    88.7 & 88.4      \\ \hline
basic+VAT   & 87.0     & 82.6      & 82.1   & 82.3     & 91.4     & 86.9      & 85.6   & 85.6     \\ \hline
LPAT+Bottom & 88.9     & 84.0      & 85.0   & 84.5     & 93.1     & 89.5      & 90.2   & 89.7     \\
LPAT+Top    & 90.0     & 85.5      & 86.1   & 85.8     &\textbf{95.6}    &\textbf{93.4}     &\textbf{92.1}      &\textbf{92.9}     \\ 
LPAT+All    &  \textbf{90.2}  &  \textbf{85.8}   & \textbf{86.7} & \textbf{86.2}      & 95.2     & 93.2      & 91.2   & 92.1     \\ \hline
\end{tabular}
\end{table*}

\begin{table*}[]
\caption{Detailed results in supervised setting}
\label{large and small}
\centering
\begin{tabular}{|p{0.09\textwidth}|p{0.043\textwidth}|p{0.043\textwidth}<{\centering}|p{0.043\textwidth}<{\centering}|p{0.043\textwidth}<{\centering}|p{0.043\textwidth}<{\centering}|p{0.043\textwidth}<{\centering}|p{0.043\textwidth}<{\centering}|p{0.043\textwidth}<{\centering}|p{0.043\textwidth}<{\centering}|p{0.043\textwidth}<{\centering}|p{0.043\textwidth}<{\centering}|p{0.043\textwidth}<{\centering}|p{0.043\textwidth}<{\centering}|}
\hline
\multicolumn{2}{|l|}{\multirow{2}{*}{}} & \multicolumn{6}{c|}{ST-1}                                                                    & \multicolumn{6}{c|}{ST-2}                                                                    \\ \cline{3-14} 
\multicolumn{2}{|l|}{}                  & \multicolumn{2}{c|}{Precision} & \multicolumn{2}{c|}{Recall} & \multicolumn{2}{c|}{Macro-F1} & \multicolumn{2}{c|}{Precision} & \multicolumn{2}{c|}{Recall} & \multicolumn{2}{c|}{Macro-F1} \\ \hline
\multicolumn{2}{|l|}{}                  & $\leq 5$      & $\leq 15$      & $\leq 5$     & $\leq 15$    & $\leq 5$      & $\leq 15$     & $\leq 5$      & $\leq 15$      & $\leq 5$     & $\leq 15$    & $\leq 5$      & $\leq 15$     \\ \hline
\multicolumn{2}{|l|}{basic+AT}          &  72.9               &   76.9          &            73.5      &     83.2 &             72.2     &         81.9       &72.1                &92.3              &59.7               &76.7                &65.7                 &83.8         \\ \hline
\multicolumn{2}{|l|}{basic+RDAT}      &   73.7             &      81.8       &    74.6              & 85.8     &      73.2            &   83.5             &     75.0           &     94.3         &        85.4       &     79.1           &    82.3             &      86.5             \\ \hline
\multicolumn{2}{|l|}{basic+VAT}        & 73.5        & 80.6        & 70.9        & 85.9       & 72.2       & 83.1                          &71.7                 &94.2                &84.7                 &73.0                &77.7       &82.3              \\ \hline
\multicolumn{2}{|l|}{LPAT+Bottom}       & 72.2         & 81.6       & 74.3         & 86.2        & 73.2   & 83.4     & 79.6         & 93.0         & 88.2         & 86.0          & 83.7          & 89.4             \\
\multicolumn{2}{|l|}{LPAT+Top}        & \textbf{74.5} & 83.1          & 72.1          & 85.3          & 73.7          & 84.2          & \textbf{89.3} & 94.3          & \textbf{88.3} & \textbf{91.6} & \textbf{88.4} & \textbf{92.5}         \\
\multicolumn{2}{|l|}{LPAT+All}       & 72.5          & \textbf{83.7} & \textbf{80.1} & \textbf{87.0} & \textbf{76.1} & \textbf{85.3} & 87.9          & \textbf{95.1} & 84.7          & 89.7        & 86.3          & 92.3           \\ \hline
\end{tabular}
\end{table*}

\subsection{Baselines and Metrics}
To demonstrate the performance improvement of the proposed approach, we compare it with the following method:
\begin{itemize}
\item \textbf{DT}: It is a decision tree method, which creates a tree-like model to predict the value of a target variable by learning decision rules inferred from the data features.
\item \textbf{RGF} \cite{Botezatu2016}:  It is a regularized greedy forests based method to predict the impending replacements of hard drives.
\item \textbf{RNN} \cite{xu2016health}: It is a recurrent neural network based model for predicting hard drive failure and giving health degrees, which treats the observed SMART attributes as time-sequence data.
\end{itemize}

Besides, we test the performance of the proposed LSTM based neural network when it is trained by different existing training methods.
\begin{itemize}
\item \textbf{basic}: It is the proposed LSTM based neural network trained without adversarial training methods. Specifically, it includes two dense (fully connected) layers, followed by a LSTM layer and a dense layer. 
\item \textbf{basic+AT} \cite{goodfellow2014explaining}: It is the basic neural network trained by an adversarial training method   which adds perturbations to inputs in the supervised setting.
\item \textbf{basic+RDAT} \cite{sankaranarayanan2017regularizing}: The basic model is trained by an adversarial training method  which adds perturbations to intermediate layers of networks in the supervised setting.
\item \textbf{basic+VAT} \cite{miyato2017virtual}: We use the training method which adds perturbations to inputs from the model distribution alone without necessarily using the label information. It can be applied to the semi-supervised setting.
\end{itemize}

For these baselines, we tune parameters on the validation set and report their best results on the testing set.

\begin{table*}[t]
\centering
\caption{Overall results in semi-supervised setting}
\label{tbl:semi}
\begin{tabular}{|p{0.09\textwidth}|p{0.0765\textwidth}<{\centering}|p{0.0765\textwidth}<{\centering}|p{0.0765\textwidth}<{\centering}|p{0.0765\textwidth}<{\centering}|p{0.0765\textwidth}<{\centering}|p{0.0765\textwidth}<{\centering}|p{0.0765\textwidth}<{\centering}|p{0.0765\textwidth}<{\centering}|}
\hline
           & \multicolumn{4}{c|}{ST-1}                                                                                                    & \multicolumn{4}{c|}{ST-2}                                                                                    \\ \cline{2-9}
           & \multicolumn{1}{c|}{Accuracy} & \multicolumn{1}{c|}{Precision} & \multicolumn{1}{c|}{Recall} & \multicolumn{1}{c|}{Macro-F1} & \multicolumn{1}{c|}{Accuracy} & \multicolumn{1}{c|}{Precision} & \multicolumn{1}{c|}{Recall} & Macro-F1      \\ \hline
basic+VAT  & 90.5                          & 86.7                           & 85.8                        & 86.2                         & 93.5                          & 90.2                           & 87.9                        & 89.0          \\ \hline
LPAT+All   & \textbf{92.6}                 & \textbf{89.3}                  & \textbf{88.7}               & \textbf{88.9}                 & \textbf{96.3}                 & \textbf{93.6}                  & \textbf{92.5}               & \textbf{93.8}
        \\ \hline \hline
basic+AT   & 86.8                          & 81.8                           & 81.3                        & 81.5                          & 91.0                          & 85.2                           & 84.5                        & 84.8          \\ \hline
basic+RDAT & 88.2                          & 83.4                           & 83.2                        & 83.3                          & 92.5                          & 88.3                           & 88.7                        & 88.4  
\\ \hline
\end{tabular}
\end{table*}


For the proposed LPAT, we design three variations: 
(1) LPAT-Bottom where the adversarial perturbations are added to the bottom two layers, (2) LPAT-Top where the adversarial perturbations are added to the LSTM layer and the top dense layer, and (3) LPAT-All where the adversarial perturbations are added to the all layers.



We measure the overall accuracy, and each class's precision, recall, and Macro-F1 score. For the basic prediction model, we set the two bottom dense layers with 128 units and the activation as None. The LSTM layer has 200 units. The final dense layer includes 3 units as we have 3 classes. The time sequence window for each sample is set to 20. We use RMSProp optimizer \cite{Rmsprop} with learning rate 0.001. The mini-batch size is 128. The epsilon $\epsilon$ is empirically set in $[0,50]$. The hyperparameter $\lambda$ is set between $[0,5]$. The training epochs are set to 210.

Our methods are implemented in Tensorflow \cite{abadi2016tensorflow}. All the experiments are trained and tested on four NVIDIA Tesla K80 GPUs. 

\subsection{Supervised Setting}
We first conduct experiments in the supervised setting where all the training data are labeled. As can be seen from Table \ref{tab:ST-1-results}, the proposed LPAT method achieves the best results on both datasets. It improves accuracy by $2.0\%$ and Macro-F1 by $2.9\%$ on ST-1 compared to the best baseline method. As LPAT can flexibly choose which layer and how many layers to add perturbations, it makes the training more robust and brings better performance on the testing set. On ST-2, it improves accuracy by $3.1\%$ and Macro-F1 by $4.5\%$ over the other baseline methods. These results indicate that the performance improvement is more significant on small datasets like ST-2. Considering that neural networks are more likely to encounter overfitting when trained by small datasets, we infer that our method shows good regularization and generalization ability to prevent overfitting.

RNN performs better than RGF on ST-1 while slightly worse than RGF on ST-2. ST-1 and ST-2 datasets are collected on different hard drive models whose SMART attributes are also different. RGF which automatically selects SMART attributes may perform better in feature obvious dataset \cite{Botezatu2016}. Models with the adversarial learning achieve better results than the basic one without the adversarial training. VAT slightly improves the performance than AT on the both datasets. basic+RDAT shows the best performance among the baseline methods only inferior to LPAT. It argues that adding perturbation to intermediate layers is an effective approach to improving robustness of neural networks.

We further investigate the impact of different perturbation patterns. For ST-1, LPAT-Bottom works better than only adding perturbation to the input layer while worse than LPAT-Top. Among the three perturbation patterns, LPAT-All achieves the best results, which could be partly attributed to the LSTM layer. Adding perturbations to the bottom layers before the LSTM layer will make perturbations wrapped in the LSTM layer and show less effect on the loss function. It undermines the effectiveness of adversary perturbations. Thus, LPAT-Bottom is less effective than adding perturbations after the LSTM layer. For ST-2, LPAT-Top achieves the best performance on accuracy and Macro-F1, though; its performance is only marginally better than LPAT-All. Based on these results, we speculate that adding perturbations to top or all layers can achieve better regularization ability.

Table \ref{large and small} details the results on each class. Note that the size of the training samples of class `1' is two times larger than that of class `0' on both datasets. Precision represents the ability for a classifier to capture hard drives fails in $5$ and $5-15$ days precisely. Recall represents the failure detection rate, i.e. the fraction of failed drives that are correctly classified as failed. The proposed methods outperform basic+AT, basic+RDAT, and basic+VAT in terms of precision and recall on both datasets. In addition, LPAT+All improves Macro-F1 by $3.9\%$ on class `0' and $2.2\%$ on class `1'  than basic+VAT for ST-1, while LPAT+Top improves Macro-F1 by $10.7\%$ on class `0' and $10.2\%$ on class `1' for ST-2. These results demonstrate that adding perturbations to different layers improves the performance of the classifier. Besides, the better performance of LPAT compared to basic+RDAT on both datasets concludes that using the KL divergence to measure perturbation could further improve robustness of neural networks. 
Based on the observation, our methods can give operators and users more flexibility to take different actions $5$ and $5-15$ days before the final failure.

\begin{figure}[tb]
\centering
\includegraphics[width=1.0\linewidth]{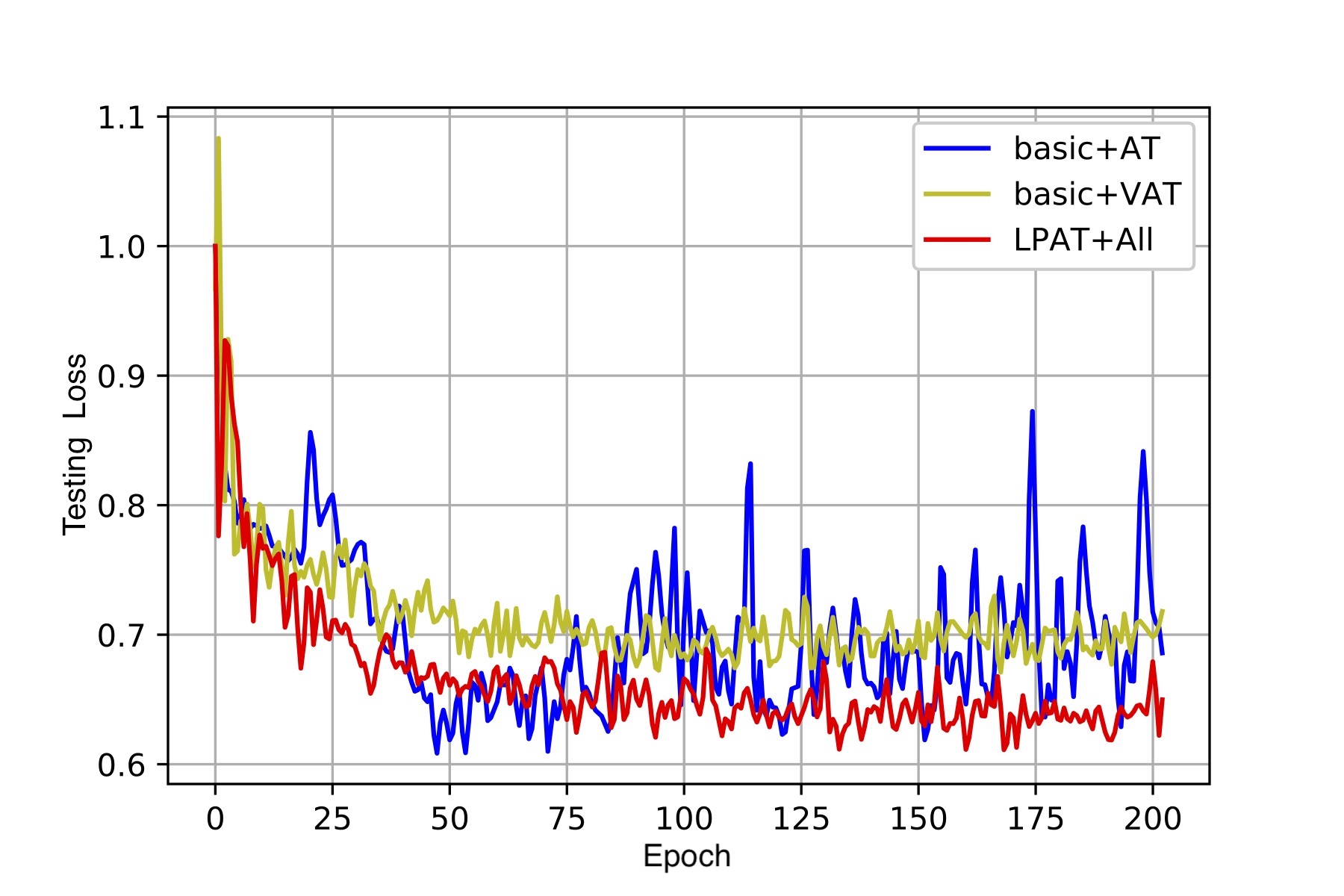}
\caption{Testing loss of AT, VAT and LPAT+All. $\lambda=1$ for all methods. $\epsilon=30$ for AT and VAT. $\epsilon=20$ on all layers for LPAT+All. The optimal value of $\epsilon$ differs between different methods. But the value of $\epsilon$ in $[0,50]$ usually performs well for above methods and provides a fair comparison.}
\label{fig:val-loss}
\end{figure}

\subsection{Semi-Supervised Setting}

In the following groups of experiments, we test the performance of the proposed method when using unlabeled data during training. Fig. \ref{fig:val-loss} plots the testing loss on ST-1 when using all unlabeled and labeled data for AT, VAT, and LPAT+All. It shows that VAT and LPAT+All which utilize the unlabeled data produce lower testing losses. The testing loss of AT becomes unstable, a sign of overfitting, with the increase of the epochs. LPAT+All generates the lowest testing loss and keeps the tendency over the training. Thanks to the resistance to the adversarial perturbations at every layer, LPAT+All performs better than VAT which merely resists to the perturbations at the input.

\begin{figure}[t]
\centering
\subfloat[Accuracy]{%
  \includegraphics[width=.251\textwidth]{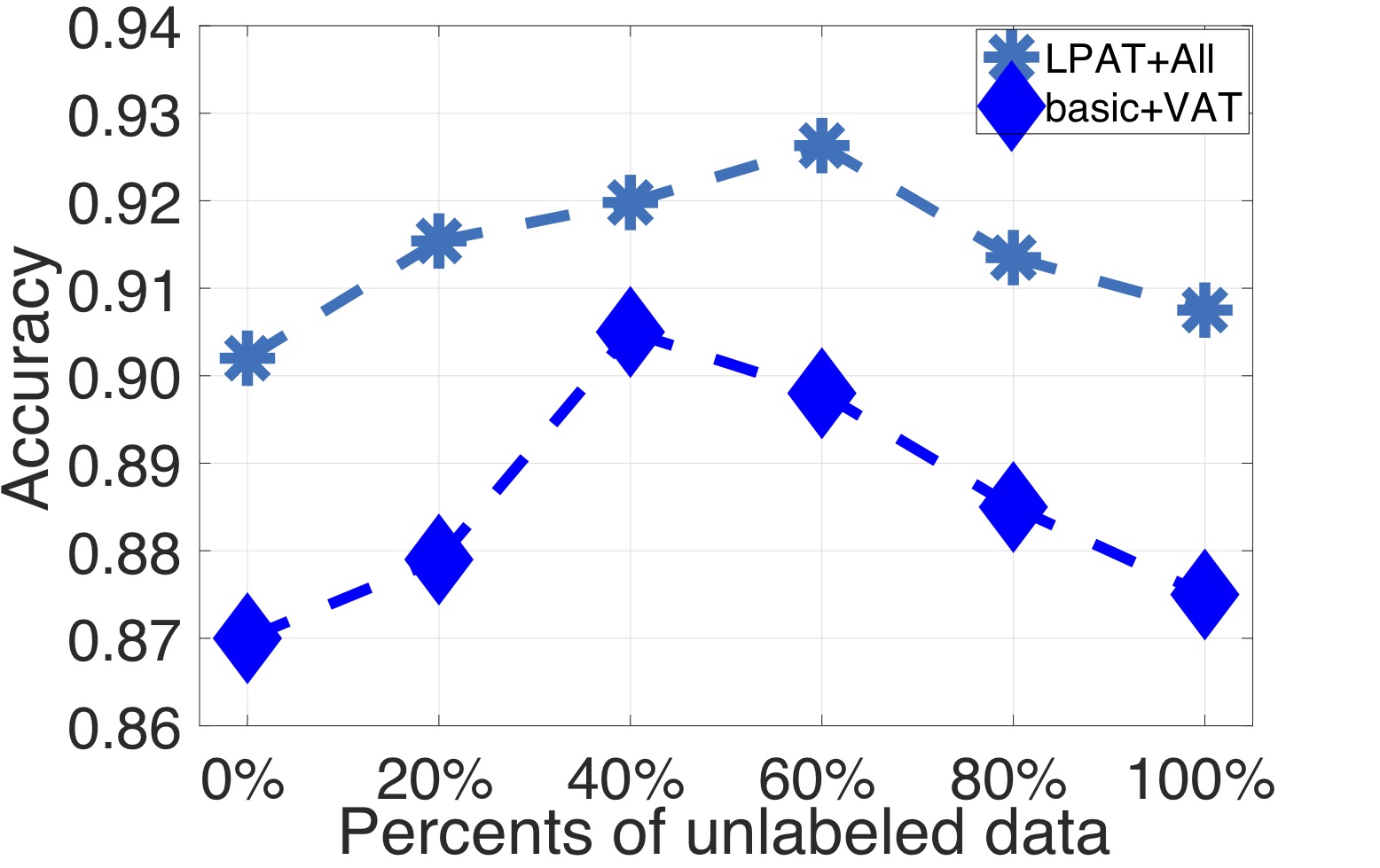}}
\subfloat[Macro-F1]{%
  \includegraphics[width=.251\textwidth]{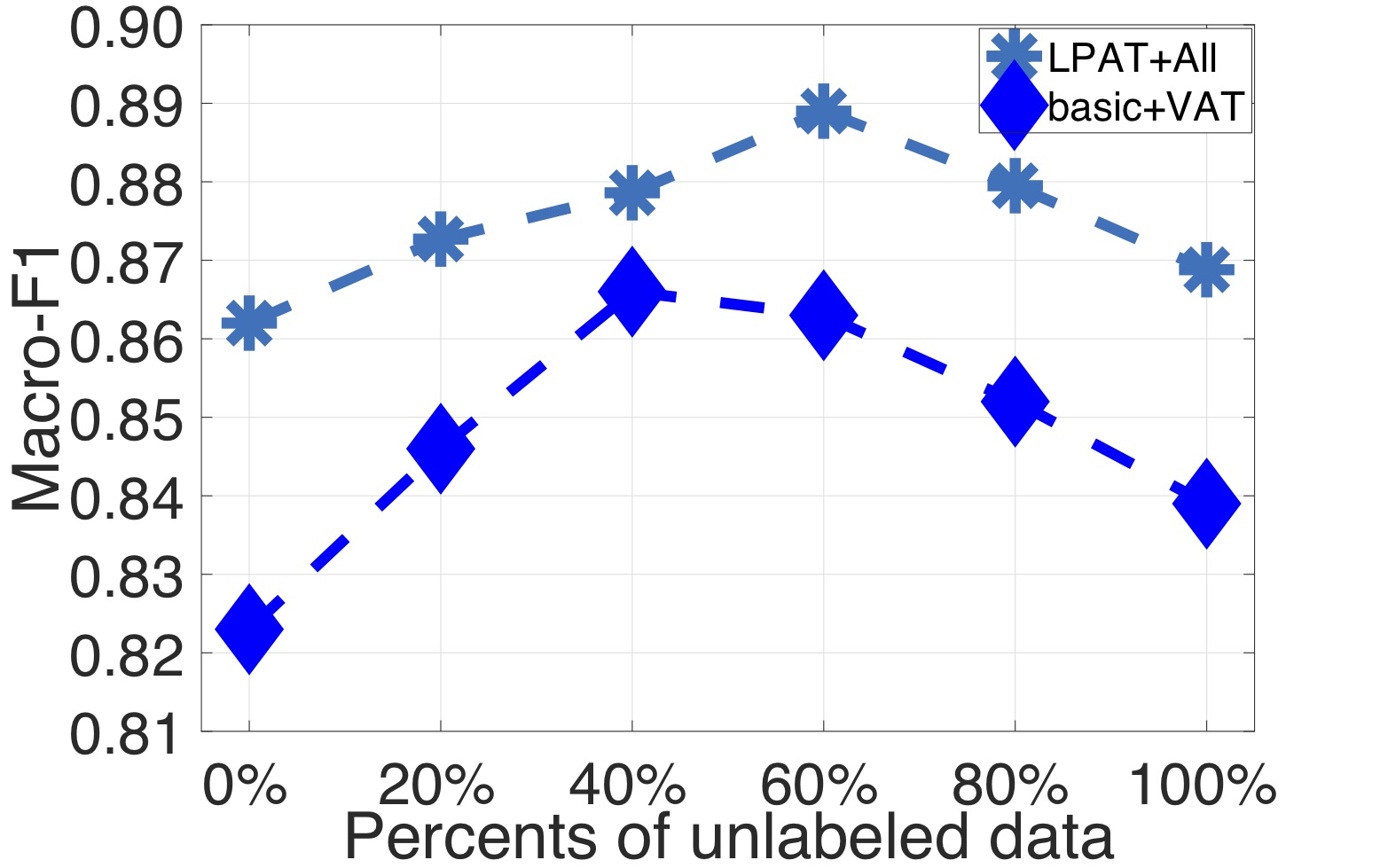}}
\caption{Accuracy and Macro-F1 on ST-1 with different amount of unlabeled data used during training. $0\%$ means only using all labeled training data; $100\%$ means using all labeled training data and all unlabeled training data.}
\label{fig:semi}
\end{figure}

Then we analyze the influence of different amounts of unlabeled data. The labeled data are always totally used during training in these experiments. But the amount of unlabeled data varies. We change the amount of unlabeled data used during training from 0\% to 100\% of all the unlabeled data. A larger value means that more unlabeled samples are used. Fig. \ref{fig:semi} gives the effect on the overall performance.

\begin{table*}[t]
\centering
\caption{Results of other tasks}
\label{tbl:new}
\begin{tabular}{|p{0.099\textwidth}|p{0.09\textwidth}<{\centering}|p{0.099\textwidth}<{\centering}|p{0.099\textwidth}<{\centering}|p{0.099\textwidth}<{\centering}|p{0.099\textwidth}<{\centering}|p{0.099\textwidth}<{\centering}|}
\hline
          & \multicolumn{2}{c|}{MNIST}    & \multicolumn{2}{c|}{ALPH.}    & \multicolumn{2}{c|}{ACCEL.}   \\ \cline{2-7}
          & Accuracy      & F1-score      & Accuracy      & F1-score      & Accuracy      & F1-score      \\ \hline
 basic+VAT  & 97.1          & 97.1          & 85.5          & 84.2          & 83.5          & 84.6          \\ \hline
 LPAT+All   & \textbf{97.8}              &    \textbf{97.7}           & \textbf{87.6} & \textbf{87.3} & \textbf{87.7} & \textbf{87.9} \\ \hline\hline
 basic+AT   & 96.1          & 96.4          & 83.2          & 82.3          & 80.5          & 80.8          \\ \hline
 basic+RDAT & 96.8 & 96.8 & 85.2          & 83.8          & 82.7          & 83.4          \\ \hline
\end{tabular}
\end{table*}

From Fig. \ref{fig:semi}, we can see that when using unlabeled data, both the accuracy and the Macro-F1 score are improved compared with the purely supervised setting where the percent of unlabeled data used is $0\%$. By extending the adversarial perturbation to the semi-supervised setting, the Macro-F1 of LPAT+All and basic+VAT are improved by $2.7\%$ and $3.9\%$, respectively. When the amount of unlabeled data increases, the accuracy and the Macro-F1 of both LPAT+All and basic+VAT firstly increase, and they achieve the highest scores around $60\%$ and $40\%$ unlabeled data, respectively. After that, the accuracy and the Macro-F1 score slightly decrease. We speculate that this is because when there are too many unlabeled data, the perturbation-based adversarial training method will focus more on resisting the perturbations to minimize the adversarial loss on unlabeled samples rather than the supervised loss. Hence, the model prioritizes being robust to perturbations rather than correctly predicting hard drives health degrees. In addition, we find that the performance of LPAT+All is almost always better than that of basic+VAT even when LPAT+All is trained without unlabeled data. It once again indicates that layerwise perturbation is a powerful approach to improving regularization and generalization of models.

Table \ref{tbl:semi} reports the best overall results of different methods. In order to clearly identify the performance improvement brought by the extension of semi-supervised setting, we list the performance of basic+AT and basic+RDAT which only use the label information to generate perturbation at the bottom of the table. For ST-1, LPAT+All improves Macro-F1 by $2.7\%$ than basic+VAT and by $5.6\%$ than basic+RDAT. For ST-2, LPAT+All improves Macro-F1 by $4.8\%$ than basic+VAT and by $5.4\%$ than basic+RDAT. This results show that the extension of semi-supervised setting could considerably improve the performance. Besides, the improvement brought by the layerwise perturbation is usually more notable on small dataset. 

\subsection{Parameter Analysis}

\begin{figure}[tb]
\centering
\subfloat[Effect of $\epsilon$]{%
\includegraphics[width=.255\textwidth]{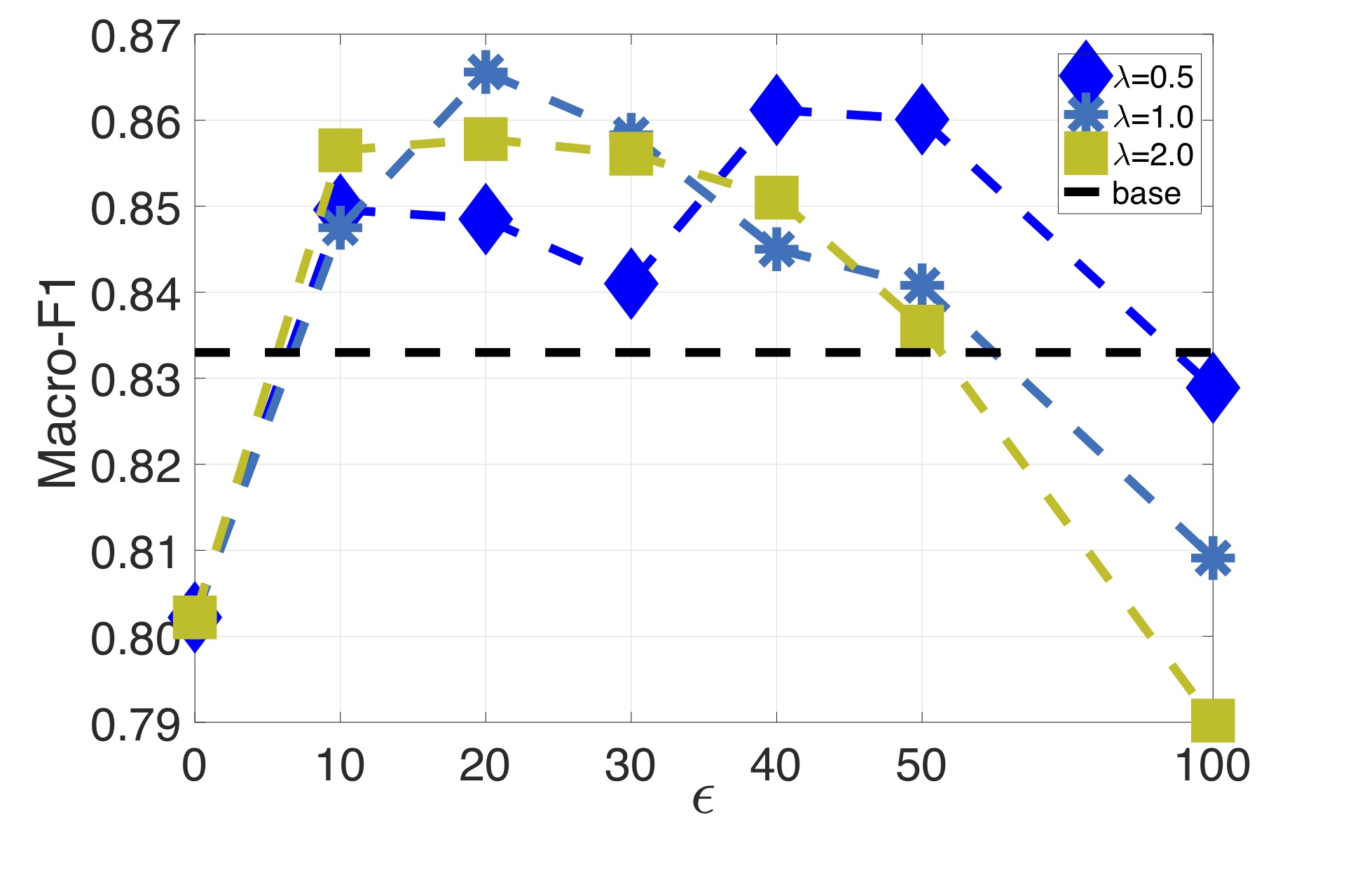}}
\subfloat[Effect of $\lambda$]{%
\includegraphics[width=.255\textwidth]{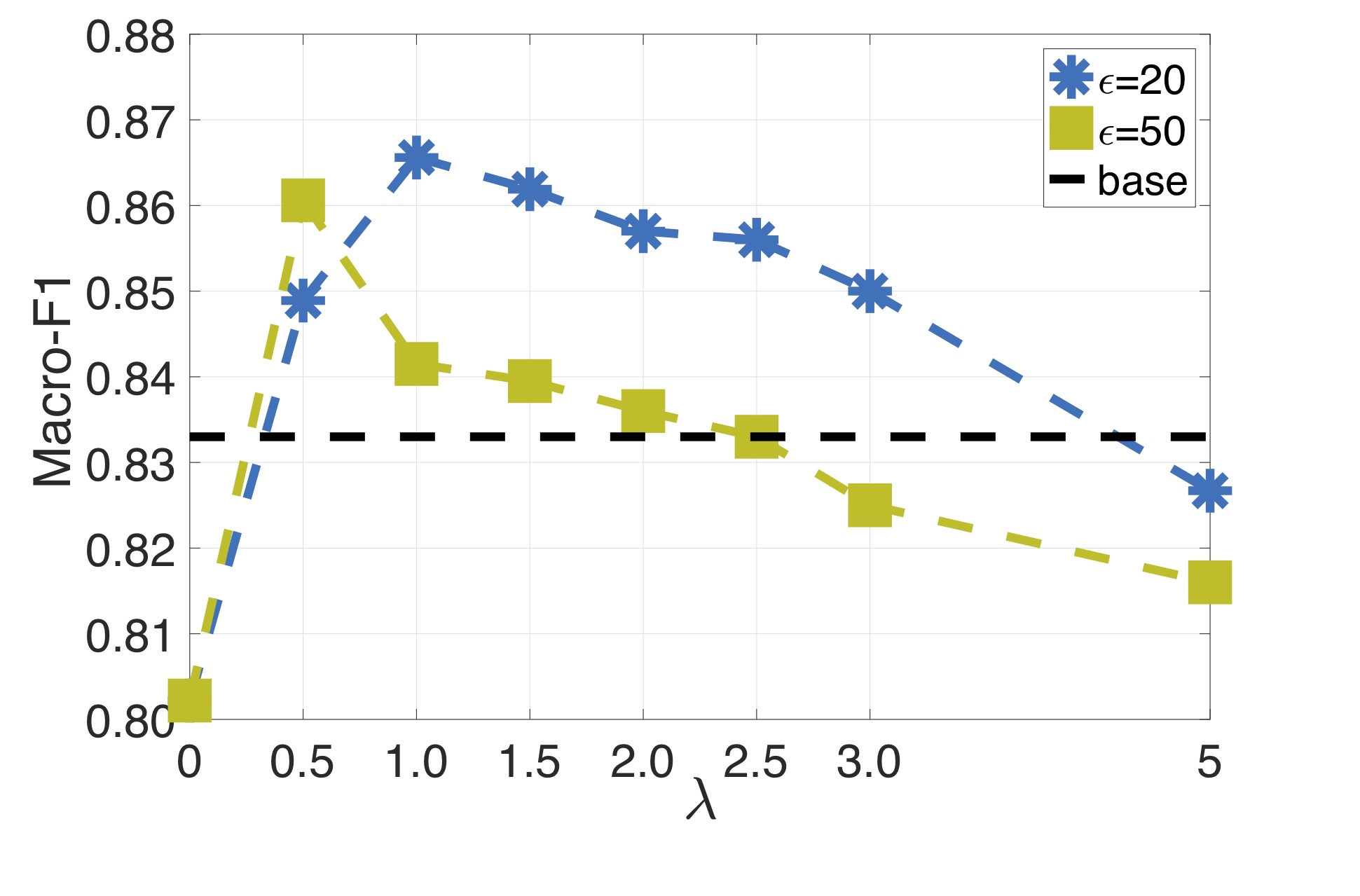}}
\caption{Effect of $\epsilon$ and $\lambda$ for supervised learning on ST-1. `base' represents the best result (basic+RDAT) of baselines.}
\label{fig:parameter-analysis}
\end{figure}

LPAT mainly involves two hyperparameters: $\epsilon$ and $\lambda$, where $\epsilon$ controls the intensity of perturbation added to different layers and $\lambda$ controls the trade-off between the supervised loss and the adversarial loss. Fig. \ref{fig:parameter-analysis}(a) shows the effects of $\epsilon$ with different values of $\lambda \in \left \{ 0.5, 1.0, 2.0 \right \}$. LPAT achieves the best result when $\epsilon=20$ and $\lambda=1$. We can find that the performance (i.e., Macro-F1) is not very sensitive to the parameter $\epsilon$. When $\epsilon \in [10,50]$, it typically can improve the neural networks performance. But when the value of $\epsilon$ is extraordinarily large, the performance would drop substantially. The perturbation is too large for neural networks to resist. Fig. \ref{fig:parameter-analysis}(b) shows the effects of $\lambda$ with different values of $\epsilon \in \left \{ 20, 50 \right \}$. The best result appears around $\epsilon=20$ and $\lambda=1$. When $\lambda>1.0$, Macro-F1 shows a descending trend. It indicates that $\lambda$ should not be too large otherwise the model would be trained to focus on the resistance against perturbation. In most cases $\lambda \in [0.5, 2.5]$, the proposed method generates better results than `base'. We find from the above results that the adversarial perturbation should be moderate to achieve the balance between the supervised loss and the adversarial loss. 

\subsection{Generality in Other Domains}
In the following experiments, we examine the generality of LPAT in other anomaly detection problems where the dataset is imbalanced. We apply the proposed methods to a image recognition task and a sequential analysis task. In the image recognition task, we construct an imbalanced dataset from MNIST. In the time series analysis task, we use a mobile application\footnote{http://www.biaffect.com} DeepMood \cite{Cao2017} to test LPAT's performance.

MNIST \cite{LeCun1998} is a large dataset of handwritten digits which includes 60,000 training images and 10,000 testing images with 10 classes from 0-9. We choose class `3' and `5' to construct the imbalanced 2-class dataset. Images of `3' and `5' are more similar to each other than other pairs, which increases the classification difficulty. In the original training dataset, there are 6,131 images of class `3' and 5,421 images of class `5'. We randomly select $5\%$ images of class `3' as labeled data and $15\%$ images as unlabeled data. All images of class `5' are used as labeled data. Then, there are 306 labeled images of class `3' and 5,421 labeled images of class `5' in the imbalanced training dataset we construct. For the testing dataset, we randomly select $20\%$ original testing images of class `3' and all original testing images of class `5' to construct the new testing dataset. For the basic model, we use a ReLU based neural network consisting of two hidden layers with the number of hidden units (1200, 1200) \cite{miyato2015distributional}.

In order to further verify the effectiveness of LPAT in real scenario, we use a mobile application DeepMood to test LPAT's performance. DeepMood harnesses the sequential information collected from the basic keystroke patterns and the accelerometer on the phone to predict the user's mood disorder. The alphanumeric character typing pattern (ALPH.) and accelerometer values pattern (ACCEL.) of 40 volunteers were collected over 8 weeks. There are 722 positive samples (disorder) and 7,500 negative samples (health) in the original dataset. We mask the labels of 30\% positive samples to construct the unlabeled data. We use 80\% of all labeled data as the training set and the rest as the testing set. The single-view DNN proposed in \cite{Cao2017} is used as the basic model.

Table \ref{tbl:new} shows the results of the image recognition task (MNIST) and the sequential analysis task (ALPH. and ACCEL.). As the two tasks are binary classification problems, we use F1-score instead of Macro-F1. The performance of four representative methods is listed. Note that basic+AT and basic+RDAT are applied to the supervised setting, and so they cannot utilize the unlabeled data. It can be found that LPAT+All demonstrates the best performance on the three datasets, which verifies the generality of our proposed methods. 

On MNIST, the accuracy and the F1-score of LPAT+All are $0.7\%$ and $0.6\%$ higher than those of basic+VAT, respectively. The slight improvement is partly due to the fact that MNIST is a relatively easy task where the accuracy and the F1-score are already quite high even without adversarial training. It is difficult to further improve the performance on MNIST.

On the two datasets of the mobile application DeepMood, LPAT+All shows the ability to substantially improve the performance. The F1-score of LPAT+All is over $3\%$ higher than the best baseline methods on both the two datasets. Basic+RDAT which adds adversarial perturbations to intermediate layers performs better than basic+AT only adding perturbations to inputs. But the performance of basic+RDAT is slightly inferior to that of basic+VAT which can be applied in the semi-supervised setting. This observation indicates that using KL divergence and unlabeled data is a little more powerful than adding perturbations to intermediate layers on these two datasets.